\documentclass[journal]{IEEEtran}

\ifCLASSINFOpdf
\else
   \usepackage[dvips]{graphicx}
\fi
\usepackage{url}

\usepackage{graphicx}
  
\usepackage{helvet}
\usepackage{courier} 
\usepackage{algorithm}
\usepackage{algorithmic}
\usepackage{csquotes} 
\usepackage{color}
\usepackage{indentfirst}
\usepackage{subfigure}  
\usepackage{cite}
\usepackage{mathrsfs}
\usepackage{textcomp,booktabs}
\usepackage{amssymb} 
\usepackage{pifont}

\usepackage{multicol}
\usepackage{lipsum}
\usepackage{cuted} 
\usepackage[group-separator={,}]{siunitx}
\usepackage{threeparttable}

\usepackage{times}
\usepackage{epsfig} 
\usepackage{booktabs}
\usepackage{courier}
\usepackage{amsmath} 
\usepackage{csquotes} 
\usepackage{color}
\usepackage{paralist}
\usepackage{amssymb}
\usepackage{indentfirst}
\usepackage{subfigure}
\usepackage{float}
\usepackage{multirow} 
\usepackage[colorlinks, linkcolor=blue, anchorcolor=blue, citecolor=blue]{hyperref}
\usepackage{tabularx}

\begin{document}

\title{Semantic-guided Pixel Sampling for Cloth-Changing Person Re-identification}

\author{Xiujun Shu, Ge Li, Xiao Wang, Weijian Ruan, Qi Tian %, \IEEEmembership{Fellow, IEEE}
	
\thanks{This work was supported by the grant of China Postdoctoral Science Foundation (No.2019M660209).(Corresponding author: Ge Li) }
\thanks{Xiujun Shu is with Peng Cheng Laboratory and Peking University, Shenzhen 518055, China (email:shuxj@pcl.ac.cn).}
\thanks{Ge Li is with the School of Electronic and Computer Engineering, Peking University, Shenzhen 518055, China (email:geli@ece.pku.edu.cn).}
\thanks{Xiao Wang and Weijian Ruan are with Peng Cheng Laboratory (e-mail: wangxiaocvpr@foxmail.com; rweij@whu.edu.cn).}
\thanks{Qi Tian is with the Huawei Cloud \& AI, Huawei Technologies, China. (e-mail: tian.qi1@huawei.com).}}

%\markboth{Journal of \LaTeX\ Class Files, Vol. 14, No. 8, August 2015}
\maketitle

\begin{abstract}
Cloth-changing person re-identification (re-ID) is a new rising research topic that aims at retrieving pedestrians whose clothes are changed. This task is quite challenging and has not been fully studied to date. Current works mainly focus on body shape or contour sketch, but they are not robust enough due to view and posture variations. The key to this task is to exploit cloth-irrelevant cues. This paper proposes a semantic-guided pixel sampling approach for the cloth-changing person re-ID task. We do not explicitly define which feature to extract but force the model to automatically learn cloth-irrelevant cues. Specifically, we firstly recognize the pedestrian's upper clothes and pants, then randomly change them by sampling pixels from other pedestrians. The changed samples retain the identity labels but exchange the pixels of clothes or pants among different pedestrians. Besides, we adopt a loss function to constrain the learned features to keep consistent before and after changes. In this way, the model is forced to learn cues that are irrelevant to upper clothes and pants. We conduct extensive experiments on the latest released PRCC dataset. Our method achieved 65.8\% on Rank1 accuracy, which outperforms previous methods with a large margin. The code is available at \textcolor{magenta}{\url{https://github.com/shuxjweb/pixel\_sampling.git}}.
\end{abstract}


\IEEEpeerreviewmaketitle

\section{Introduction}

\IEEEPARstart{P}{erson} re-identification (re-ID) plays important roles in surveillance systems and has achieved great progress in recent years \cite{Zheng2016SIFT, karanam2019a, leng2019a, ye21reidsurvey}. However, previous methods mainly focus on short-term re-ID, assuming that pedestrians would not change their clothes \cite{2015Similarity, sun2018beyond, 2020AsNet, chen2020pseudo}. This assumption limits the application of re-ID in real scenarios. For example, people may wear different clothes on different days, and a criminal can deliberately change his clothes to mislead the surveillance system. It is challenging for existing re-ID systems that mainly utilize clothing cues to recognize targets in these scenarios.

Recently, some scholars in this field have made some efforts on cloth-changing settings. Yang et al. \cite{2019Person} transferred the body contour sketch to a polar coordinate space. This method has a strong assumption that a person only changes his clothes moderately. Wan et al. \cite{wan2020person} provided a preliminary solution that jointly utilizes holistic body and head images. Yu et al. \cite{yu2020cocas} defined an easier cloth-changing re-ID setting that the query contains an image and a clothes template. Fan et al. \cite{fan2020learning} utilized radio signals for long-term person re-ID. But it is difficult to deploy large scale wireless devices. Gait, the walking pattern of individuals, has been studied extensively \cite{2019Gait, 2019EV_Gait, 2020Gait_Chao}. Recently, it is applied to the cloth-changing setting \cite{2021Cloth_Changing}, but its performance is still poor and only used as an auxiliary cue. To sum up, all the methods aim at exploiting cloth-irrelevant cues, but no methods are effective enough. This task has not been fully studied to date. 

\begin{figure}[t]
	\centering
	\includegraphics[width=\linewidth]{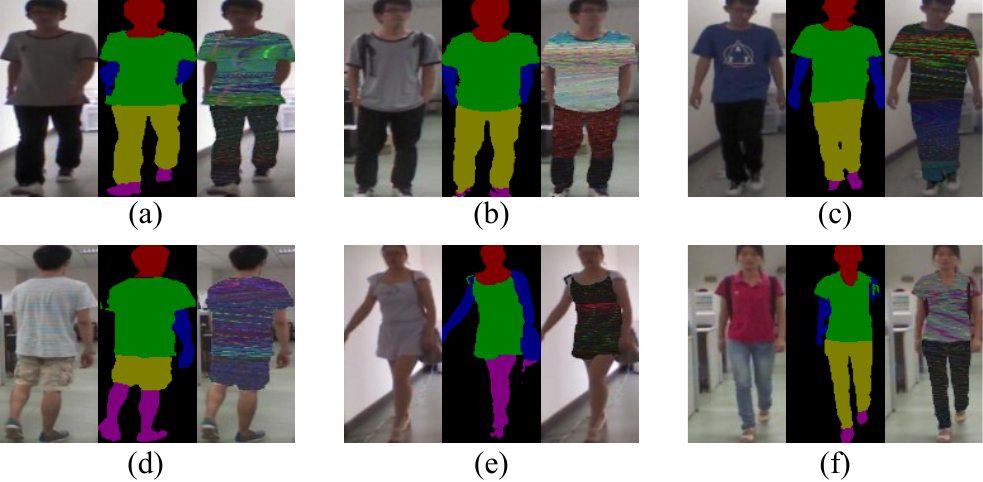} 
	\caption{\textbf{Visualization of samples changed with semantic-guided pixel sampling.}
		(a), (b) and (c) have the same identity, and the clothes of (a) and (b) are the same but different from (c). (d), (e) and (f) are different people. The three images represent the initial one, the segmentation parts, and the generated one with pixel sampling, respectively.
	}
	\label{fig:motivation}
\end{figure}

In this paper, we investigate the issue of cloth-changing person re-ID. Unlike existing methods that use some pre-defined cues, {\itshape e.g.,} body shape, radio signals, or gait, we still utilize the RGB images due to the wide availability. We propose a simple but effective method that does not explicitly define which feature to extract but forces the model to automatically learn cloth-irrelevant cues. Specifically, we first utilize a pre-trained human parsing model \cite{zhang2019freeanchor} to get body parts, then change the clothes by sampling pixels from other pedestrians. As shown in Fig.~\ref{fig:motivation}, we change the upper clothes or pants but retain the pixels of other parts. By constraining the learned features before and after changing, the model is forced to learn cloth-irrelevant cues.

\begin{figure*}[t]
	\centering
	\includegraphics[height=6.7cm]{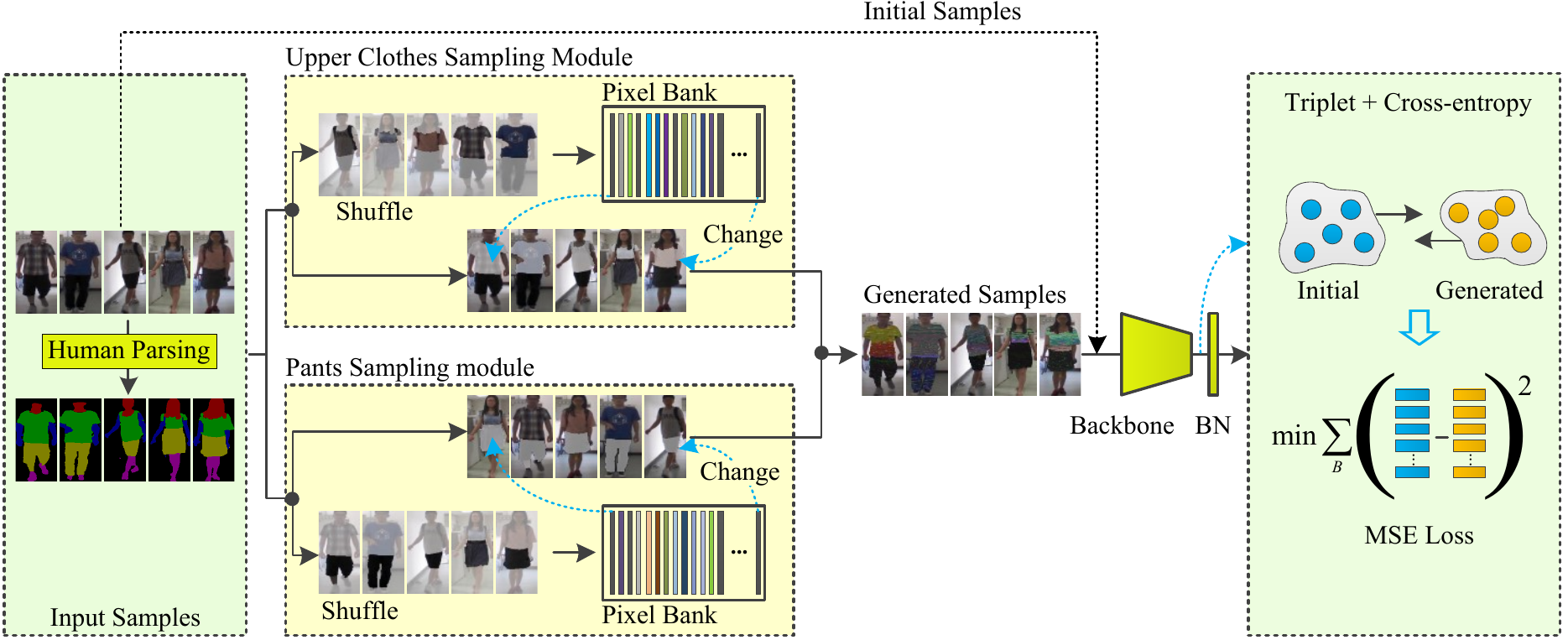} 
	\caption{\textbf{The illustration of the proposed semantic-guided pixel sampling approach.}
		The pixel sampling architecture includes two sub-modules: upper clothes sampling and pants sampling. The generated and initial samples are both used to train the backbone network. Besides triplet and cross-entropy losses, the MSE loss is utilized to constrain the learned features.
	}
	\label{fig:framework}
\end{figure*}

Our contributions can be summarized as follows: 1) A semantic-guided pixel sampling approach is proposed for cloth-changing person re-identification. Besides, a loss function is adopted to constrain the learned features. This approach could learn cloth-irrelevant cues and is suitable for cloth-changing settings. 2) We conduct extensive experiments on the latest released dataset. Experimental results show that our approach outperforms previous methods with a large margin.

\section{Proposed Method}
The key contribution is the semantic-guided pixel sampling that forces the model to learn cloth-irrelevant cues. As shown in Fig.~\ref{fig:framework}, the pixel sampling includes two sub-modules: upper clothes sampling and pants sampling. The generated and initial samples are both used to train the backbone network. More details can be seen in the following sections.

\subsection{Semantic-guided Pixel Sampling}
For the input samples, we firstly utilize a pre-trained human parsing model \cite{li2020self, liang2015deep} to get body parts. Since the model predicted 18 parts, we recombined them to get six parts: background, head, upper-clothes, pants, arms, and legs. In our approach, we only consider the scenario that the upper-clothes and pants are changed.

The framework includes two sub-modules: upper clothes sampling and pants sampling. The process of the two sub-modules is similar. Let us take the sub-module ``upper clothes sampling'' as an example. Assume that the input mini-batch samples are denoted as $\mathbb{X}=[x_1, x_2, ...,x_B]$, where $B$ is the mini-batch size. 
$x_i$ is an image with a shape of $C\times H\times W$, which denote the numbers of channel, height, and width, respectively. 
The semantic segmented results are denoted as $\mathbb{S}=[s_1, s_2, ...,s_B]$. $s_i$ has a shape of $1\times H\times W$. The pixel values in $s_i$ belong to $\{0, 1, 2, 3, 4, 5\}$, which represent the six parts, respectively. 

First, we randomly shuffle $\mathbb{X}$ to get $\bar{\mathbb{X}}=[x_{a_1}, x_{a_2}, ...,x_{a_B}]$, where $a_i$ is the new subscript, $a_i\in [1,B]$. The semantic segmented results $\mathbb{S}$ is changed as $\bar{\mathbb{S}}=[s_{a_1}, s_{a_2}, ...,s_{a_B}]$. Since each pixel of $x_{a_i}$ can be represented as a vector $\bar{v}_j$ with $C$ values, the input sample $x_{a_i}$ has $W\cdot H$ pixel vectors in total. Then $\bar{\mathbb{S}}$ is used to obtain the pixel vectors of upper clothes. All the pixel vectors of upper clothes in each mini-batch samples are stored as follows:
\begin{equation}   
\bar{\mathbb{B}}_{upper}=\{\bar{v}_j|\bar{v}_j=x_{a_i}[s_{a_i}==2], j\in[1, N_u]\},
\end{equation}
where $N_u$ is the pixel vector quantity of upper clothes in the mini-batch samples. It usually changes in different iterations. $j$ is the index of the $j^{th}$ pixel vector. ``=='' denotes equal to, and 2 is the index of upper-clothes. $\bar{v}_j$ is a vector with three (R, G, B) values. $x_{a_i}\in \bar{\mathbb{X}}$ and $s_{a_i}\in \bar{\mathbb{S}}$.

Assume all pixel vectors in $\mathbb{X}$ are denoted as $\mathbb{V}_\mathbb{X}=[v_1, v_2, ...,v_{m_1},...,v_{m_{N_u}},..., v_{M}]$, in which $[v_{m_1},..,v_{m_{N_u}}]$ belong to upper clothes and $M$ is the total number of pixels, $M=B\cdot H\cdot W$. 
Next, $\bar{\mathbb{B}}_{upper}$ is used to change the pixels of upper clothes in $\mathbb{V}_\mathbb{X}$. The changed pixel vector can be represented as:
\begin{equation}
\mathbb{V}_\mathbb{X}=[v_1, v_2, ...,\bar{v}_1,...,\bar{v}_{N_u},..., v_{M}],
\end{equation} 
where $\bar{v}_1,...,\bar{v}_{N_u}\in \bar{\mathbb{B}}_{upper}$.
 
Similarly, the sub-module of pants sampling is operated in the same way. Assume that the index of pants is 3 and the pixel bank is denoted as $\bar{\mathbb{B}}_{pants}$, the pixels of pants in $\mathbb{V}_\mathbb{X}$ are changed by $\bar{\mathbb{B}}_{pants}$. Finally, the generated and initial samples are both used to train the model. 

We can see that the semantic-guided pixel sampling is simple but effective. By changing the pixels of upper clothes and pants, it could force the model to learn more cloth-irrelevant cues. Its effectiveness has been verified by conducting extensive experiments. 
 
\subsection{Loss Function}
The upper clothes and pants of input samples are randomly changed through the above operations. The generated samples retain identity information and are used as training samples together with the initial ones. Since the upper clothes and pants occupy a large proportion of pixels, the generated samples and initial ones have different appearances. To force the model to focus on cloth-irrelevant cues, we adopt the mean square error to constrain the learned features.
\begin{equation}
\mathcal{L}_{mse} = \frac{1}{B}\sum_{i=1}^{B}\Big(||f_i-f_i^{'}||_2\Big), 
\end{equation}
where $||\cdot||_2$ denotes the $L_2$ norm. $f_i$ is the $i^{th}$ feature of $\mathbb{X}$, and $f_i^{'}$ is the feature after changing upper clothes or pants.

Besides $\mathcal{L}_{mse}$, the cross-entropy and triplet losses also need to be used to learn discriminative features.
\begin{align}   
&\mathcal{L}_{i} = \frac{1}{2B}\sum_{i=1}^{2B}\Big(-y_i\cdot\log p(x_i)\Big),\\
&\mathcal{L}_{t} = \frac{1}{2B}\sum_{i=1}^{2B}\Big(max\{m+d_{max}^{p}(f_i)-d_{min}^{n}(f_i), 0\}\Big), 
\end{align} 
where $p_i$ and $y_i$ denote the predicted probability and its label, respectively. $m$ is set as 0.3. $d_{max}^p$ and $d_{min}^n$ denote the maximum distance of positive sample pairs and minimum distance of negative sample pairs. The mini-batch size is $2B$, because the initial and generated samples are both utilized as training samples. The total loss can be denoted as:
\begin{equation}
\mathcal{L} = \mathcal{L}_{mse} + \mathcal{L}_{i} + \mathcal{L}_{t}. 
\end{equation}

\section{Experiments} 

\subsection{Datasets and Settings}

\subsubsection{Datasets}
We evaluate our method on the latest released Person Re-id under moderate Clothing Change (PRCC) dataset \cite{2019Person}. It was collected for the task of cloth-changing person re-ID. PRCC consists of 221 identities with three camera views. The training set contains 150 identities with 17,896 images. The gallery set contains 71 identities with 3,384 images. This dataset has two queries: same clothes and cross-clothes. The query with the same clothes has 71 identities with 3543 images, and the cross one has 71 identities with 3873 images.

\subsubsection{Implementation Details}
ResNet50 \cite{he2016deep} is used as the backbone network, which is initialized with ImageNet \cite{russakovsky2015imagenet} pre-trained model. The SGD optimizer is used in our experiments. The images are resized to 256$\times$128. The random cropping and horizontal flipping are utilized as the augmentation methods. The initial learning rate is set as $3.5\times 10^{-3}$. The mini-batch size is 64. First, we randomly sample 16 identities from the training dataset and then randomly sample 4 instances for each identity. For fair comparison, this sampling strategy was used in all the following experiments.

\subsection{Comparison With the State-of-the-Art} 
As shown in Table~\ref{table:compare_sota}, the experiments are divided into two groups: cross-clothes and same clothes. ``Cross-clothes'' means that persons wear different clothes between query and gallery. ``Same clothes'' means that they wear the same clothes. Since PRCC is newly released, there are few methods evaluated on it so far. In the setting of same clothes, conventional RGB-based methods, {\itshape e.g.,} PCB, MGN, and HPM, work well and achieve high accuracy. For example, PCB \cite{sun2018beyond} achieves 99.8\% on Rank1 and 97.0\% on mAP. This is much higher than sketch-based and gait-based methods. Because clothes, as the major cues, have the strong discriminative ability. However, sketch-based and gait-based methods mainly focus on cloth-irrelevant cues. The lack of clothing cues leads to the loss of final performance. In the setting of cross-clothes, clothing cues are no longer reliable. Conventional RGB-based methods have a large performance drop. Recently, gait recognition has been applied to the cloth-changing setting. GI-ReID \cite{2021Cloth_Changing} utilizes gait to enhance RGB cues and achieves 37.6\% on Rank1 accuracy. However, it concluded that directly using the gait fails to get satisfactory results. SPT\cite{2019Person} transfers the body contour sketch into a polar coordinate space. Although designed for the cloth-changing setting, it only achieves 34.4\% on Rank1 accuracy. This is because it only uses the body contour sketch and ignores the appearance cues. Our method achieves 65.8\% on Rank1 and 61.2\% on mAP, which are much better than other methods. It demonstrates the effectiveness of our method for cloth-changing person re-ID setting.

\begin{table}[!t]
	\centering
	%\scriptsize
	\renewcommand\arraystretch{1.2}
	\caption{\textbf{Performance comparison of cross-clothes and same clothes on the PRCC dataset}.}
	%\scalebox{0.98}[0.98]{  
	%\resizebox{\linewidth}{!}{ 
	\begin{tabular}{p{1.7cm}|p{1.5cm}<{\centering}|p{0.67cm}<{\centering}p{0.67cm}<{\centering}|p{0.67cm}<{\centering}p{0.67cm}<{\centering}}
		\hline  
		\multirow{2}{*}{\textbf{Methods}} & \multirow{2}{*}{\textbf{Type}} & \multicolumn{2}{c|}{\textbf{Cross-clothes}} & \multicolumn{2}{c}{\textbf{Same clothes}}  \\ 
		
		\cline{3-6} 
		& &\textbf{R1}  & \textbf{mAP}  & \textbf{R1}  & \textbf{mAP} \\ 
		\hline
		HA-CNN\cite{li2018harmonious}   &  & 21.8 &- &82.5 &- \\
		
		STN\cite{Zhong2017Re}   &  & 27.5 &- &59.2 &- \\
		
		PCB\cite{sun2018beyond}   & RGB & 41.8 &38.7 &\textbf{99.8} &97.0 \\

		MGN\cite{wang2018learning}   &  & 33.8 &35.9 &99.5 &\textbf{98.4} \\ 
		
		HPM\cite{fu2019horizontal}  &  & 40.4 &37.2 &99.4 &96.9 \\ 
		
		\hline 
		GI-ReID\cite{2021Cloth_Changing}	& Gait+RGB  & 37.6 &- &86.0 &- \\

		\hline
		VGG16\cite{2015Karen}	&   & 18.79 &- &54.00 &- \\ 
		SketchNet\cite{wu2018exploit}  & Sketch  & 17.9 &- &64.6 &- \\ 
		
		SPT\cite{2019Person}  &   & 34.4 &- &64.2 &- \\

		\hline
		Ours     & RGB & \textbf{65.8} &\textbf{61.2} &99.5 &96.7  \\ 
		
		\hline
	\end{tabular}\label{table:compare_sota}
	%}
\end{table}

\begin{table}[!t]
	\centering
	%	\scriptsize
	\renewcommand\arraystretch{1.2}
	\caption{\textbf{Ablation studies of components in our framework} PS: Pixel Sampling, RE: Random Erasing, MSE: MSE Loss}.  
	%\scalebox{0.9}[0.9]{
	%	\resizebox{\linewidth}{!}{ 
	\begin{tabular}{p{0.97cm}<{\centering}|p{0.97cm}<{\centering}|p{0.97cm}<{\centering}|p{0.97cm}<{\centering}|p{0.97cm}<{\centering}|p{0.97cm}<{\centering}}
		\hline
		\textbf{Baseline} 	& \textbf{PS}   & \textbf{MSE}   & \textbf{RE}	& \textbf{R1}   	& \textbf{mAP} \\ 
		\hline 
		$\checkmark$    & &	&	
		& 36.3  & 33.7 \\
		
		$\checkmark$    & $\checkmark$ & &	
		& 62.2  & 56.6 \\

		$\checkmark$    & $\checkmark$ & $\checkmark$ &
		& 65.8  & 61.2 \\
		
		$\checkmark$    &  & & $\checkmark$	
		& 46.5  & 45.3 \\
		
		$\checkmark$    & $\checkmark$ & $\checkmark$ & $\checkmark$
		& 67.7  & 62.3 \\
		\hline
	\end{tabular}\label{table:ablation_eraser}
\end{table}

\begin{table}[!t]
	\centering
	%\scriptsize
	\renewcommand\arraystretch{1.2}
	\caption{\textbf{The effectiveness of pixel sampling on conventional methods}.w/o: without, +: with}
	%\scalebox{0.98}[0.98]{  
	%\resizebox{\linewidth}{!}{ 
	\begin{tabular}{p{2cm}|p{1.08cm}<{\centering}p{1.08cm}<{\centering}|p{1.08cm}<{\centering}p{1.08cm}<{\centering}}
		\hline  
		\multirow{2}{*}{\textbf{Methods}} & \multicolumn{2}{c|}{\textbf{w/o pixel sampling}} & \multicolumn{2}{c}{\textbf{+pixel sampling}}  \\ 
		
		\cline{2-5} 
		& \textbf{R1}  & \textbf{mAP}  & \textbf{R1}  & \textbf{mAP} \\ 
		\hline
		
		PCB\cite{sun2018beyond}   & 41.8 &38.7 &55.3 &50.6 \\

		MGN\cite{wang2018learning}   &  33.8 &35.9 &59.5 &56.5 \\ 
		
		HPM\cite{fu2019horizontal}  &  40.4 &37.2 &56.5 &51.5 \\ 
		
		\hline
	\end{tabular}\label{table:compare_pix}
	%}
\end{table}

\begin{table}[!t]
	\centering
	%\scriptsize
	\renewcommand\arraystretch{1.2}
	\caption{\textbf{The effect of head and body cues}. The codes of Head and Head+Body are based on BoT\cite{luo2019bag}}
	%\scalebox{0.98}[0.98]{  
	%\resizebox{\linewidth}{!}{ 
	\begin{tabular}{p{2cm}|p{1.08cm}<{\centering}p{1.08cm}<{\centering}|p{1.08cm}<{\centering}p{1.08cm}<{\centering}}
		\hline  
		\multirow{2}{*}{\textbf{Methods}} &  \multicolumn{2}{c|}{\textbf{Cross-clothes}} & \multicolumn{2}{c}{\textbf{Same clothes}}  \\ 
		
		\cline{2-5} 
		&\textbf{R1}  & \textbf{mAP}  & \textbf{R1}  & \textbf{mAP} \\ 
		\hline
		
		Face\cite{2016A}	   & 2.97 &- &4.75 &- \\  
		
		Head	   & 36.8 &27.5 &88.2 &67.7 \\  
		Head+Body  & 45.4 &38.1 &99.8 &93.7 \\ 
		
		\hline
		Ours      & 65.8 &61.2 &99.5 &96.7  \\ 
		
		\hline
	\end{tabular}\label{table:compare_head}
	%}
\end{table}

\begin{figure}[t]
	\centering
	\includegraphics[width=\linewidth,height=4.3cm]{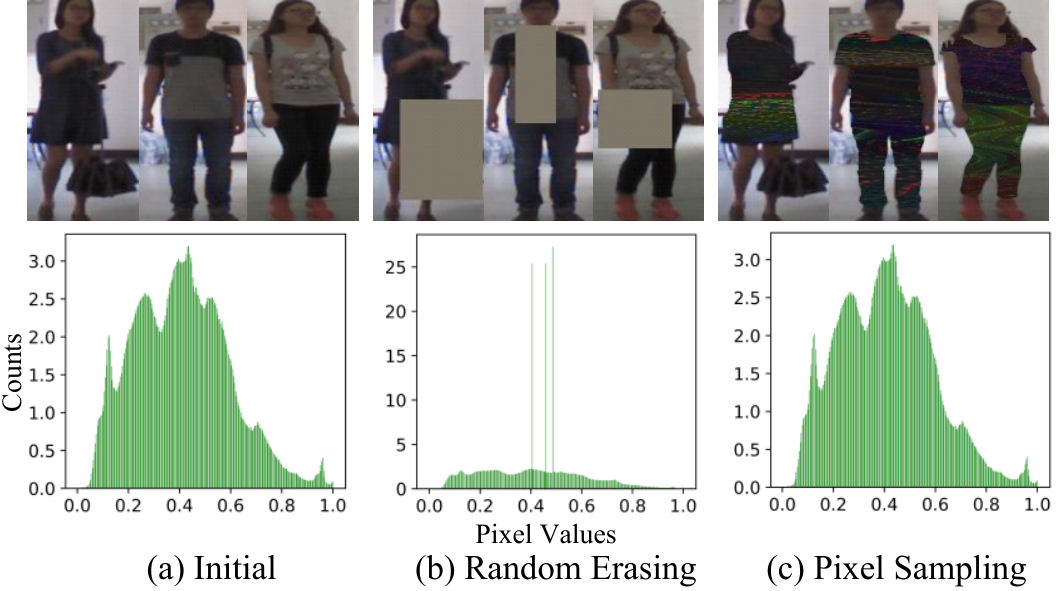} 
	\caption{\textbf{Comparison between random erasing and our method.}
	}
	\label{fig:random_erasing}
\end{figure}

\subsection{Ablation Study}
\subsubsection{Effectiveness of Components}
In our method, the baseline is based on BoT \cite{luo2019bag}. As shown in Table~\ref{table:ablation_eraser}, the baseline achieves 36.3\% on Rank1 accuracy. When integrated with the pixel sampling modules, the performance is improved to 62.2\%. This is a significant improvement. With the help of MSE loss, the performance is further boosted to 65.8\% on Rank1. This is because that the MSE Loss further constrains the model to learn cloth-irrelevant cues. Besides, we also applied pixel sampling to conventional methods in Table~\ref{table:compare_pix}. It shows us that the performances of the three methods are all boosted with large margins. The above results fully demonstrate the important roles of pixel sampling and MSE loss for cloth-changing settings.

\subsubsection{Comparison with Random Erasing}
Random erasing \cite{zhong2020random} has been widely used in the re-ID community. 
%It randomly selects a rectangle region in an image and erases its pixels with random values. This operation could boost the performance to some extent. 
As shown in Table~\ref{table:ablation_eraser}, it achieves 46.5\% on Rank1 accuracy, but still poorer than ours. With the help of random erasing, our method further boosts the performance to 67.7\%. Fig.~\ref{fig:random_erasing} shows that random erasing has several specific values, resulting the changes of distributions. Different distributions between the training and testing sets would decay the final performance.

\subsection{Visualization}
\begin{figure}[t]
	\centering 
	\includegraphics[width=\linewidth,height=6.94cm]{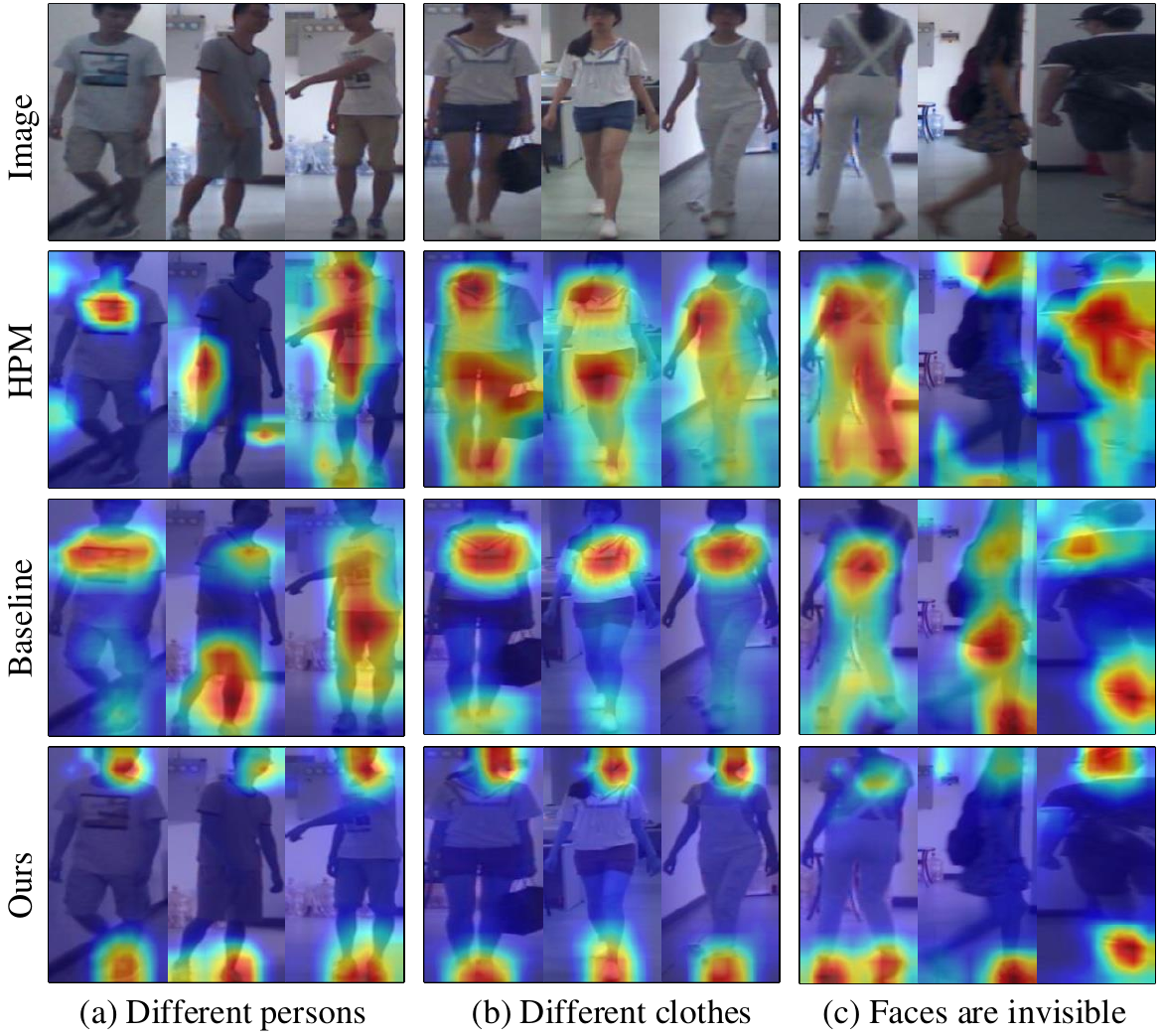}
	\caption{\textbf{Visualization of heat maps.}
		The baseline and HPM focus on the whole body, especially the upper clothes and pants. Our method focus on cloth-irrelevant cues, {\itshape e.g.,} head, legs, and shoes.
	}
	\label{fig:attention}
\end{figure}

To verify which cues the model has learned, we visualize the heat maps in Fig.~\ref{fig:attention}. The class activation mapping (CAM) \cite{zhou2016learning} is used to generate the heat maps. The brighter the pixels are, the more attention the model pays to. It shows us that the baseline and HPM focus on the whole body, especially the clothes. They pay little attention to the heads. This is due to the low proportion of pixels in the head. Clothing cues are no longer reliable in the cloth-changing setting. This leads to the poor performance of conventional methods. However, our method mainly focuses on the head and shoes, especially faces. Fig.~\ref{fig:attention}(c) shows some samples in which faces are invisible. We can see that hair and shoes are important cues in our method. The above visualizations show us that our method has indeed learned some cloth-irrelevant cues.

\begin{figure}[t]
	\centering
	\includegraphics[width=\linewidth, height=3.3cm]{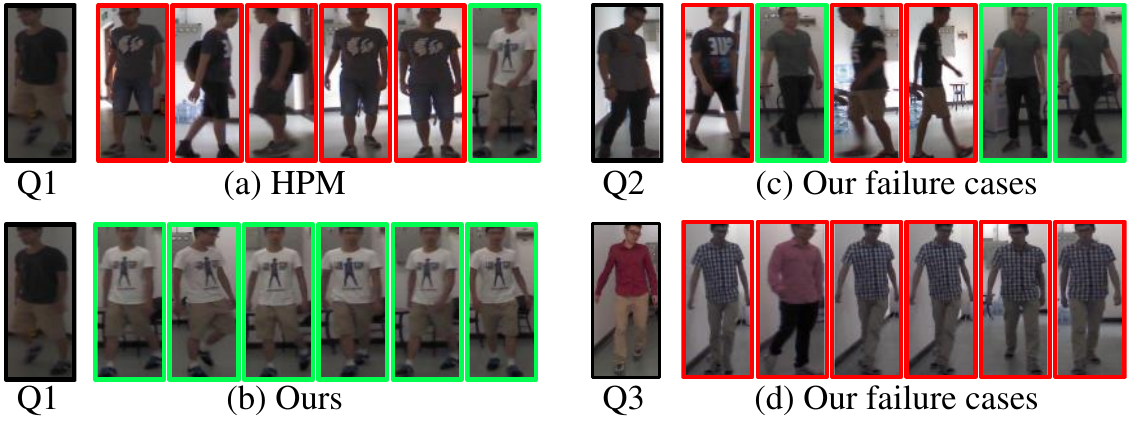}
	\caption{\textbf{The top six retrieved results.} Q denotes the query image. The red color denotes failure cases, and the green colors denotes right cases.
	}
	\label{fig:visual_rank}
\end{figure}
 
\subsection{Limitations and Discussion} 
We are still interested in whether satisfied performance can be achieved with only the head images. As shown in Table~\ref{table:compare_head}, the head achieves 36.8\% on Rank1, which is better than the face-based method. The utilization of the whole body further boosts the performance of the head, but is still poorer than our method. This demonstrates the effectiveness of our method in cloth-changing settings. In Fig.~\ref{fig:visual_rank}, we show some retrieved results. Comparing (a) and (b), the retrieved samples of HPM have similar clothes with the query, but our method could retrieve the right ones. We also show two failure cases in (c) and (d). We could see that the head cues of retrieved samples and the query are similar. Their faces are blurred. They all wear glasses and have short hairs. This indicates that our method lacks the ability to capture fine-grained head cues.

\section{Conclusion}
In this paper, we propose a semantic-guided pixel sampling approach for cloth-changing person re-identification. It changes the upper clothes or pants by sampling pixels from other pedestrians. This approach does not pre-define which features to extract, but forces the model to automatically exploit cloth-irrelevant cues, {e.g.,} head, legs. Experiments and visualization demonstrate the effectiveness of the proposed approach in cloth-changing settings. We also noticed that our method still lacks the ability of extracting fine-grained cues. Cloth-changing person re-ID is still a challenging but important task. We hope that this work could provide some reference values for future research.

%\section*{References}

\bibliographystyle{IEEEtran} 
\bibliography{ref_reid}

\begin{thebibliography}{10}
\providecommand{\url}[1]{#1}
\csname url@samestyle\endcsname
\providecommand{\newblock}{\relax}
\providecommand{\bibinfo}[2]{#2}
\providecommand{\BIBentrySTDinterwordspacing}{\spaceskip=0pt\relax}
\providecommand{\BIBentryALTinterwordstretchfactor}{4}
\providecommand{\BIBentryALTinterwordspacing}{\spaceskip=\fontdimen2\font plus
\BIBentryALTinterwordstretchfactor\fontdimen3\font minus
  \fontdimen4\font\relax}
\providecommand{\BIBforeignlanguage}[2]{{%
\expandafter\ifx\csname l@#1\endcsname\relax
\typeout{** WARNING: IEEEtran.bst: No hyphenation pattern has been}%
\typeout{** loaded for the language `#1'. Using the pattern for}%
\typeout{** the default language instead.}%
\else
\language=\csname l@#1\endcsname
\fi
#2}}
\providecommand{\BIBdecl}{\relax}
\BIBdecl

\bibitem{Zheng2016SIFT}
L.~Zheng, Y.~Yang, and Q.~Tian, ``Sift meets cnn: A decade survey of instance
  retrieval,'' \emph{IEEE Transactions on Pattern Analysis and Machine
  Intelligence}, vol.~40, no.~5, pp. 1224--1244, 2016.

\bibitem{karanam2019a}
S.~Karanam, M.~Gou, Z.~Wu, A.~Ratesborras, O.~Camps, and R.~J. Radke, ``A
  systematic evaluation and benchmark for person re-identification: Features,
  metrics, and datasets,'' \emph{IEEE Transactions on Pattern Analysis and
  Machine Intelligence}, vol.~41, no.~3, pp. 523--536, 2019.

\bibitem{leng2019a}
Q.~Leng, M.~Ye, and Q.~Tian, ``A survey of open-world person
  re-identification,'' \emph{IEEE Transactions on Circuits and Systems for
  Video Technology}, vol.~30, no.~4, pp. 1092--1108, 2020.

\bibitem{ye21reidsurvey}
M.~Ye, J.~Shen, G.~Lin, T.~Xiang, L.~Shao, and S.~C.~H. Hoi, ``Deep learning
  for person re-identification: A survey and outlook,'' \emph{IEEE Transactions
  on Pattern Analysis and Machine Intelligence}, 2021.

\bibitem{2015Similarity}
J.~Wang, N.~Sang, Z.~Wang, and C.~Gao, ``Similarity learning with top-heavy
  ranking loss for person re-identification,'' \emph{IEEE Signal Processing
  Letters}, vol.~23, no.~1, pp. 84--88, 2015.

\bibitem{sun2018beyond}
Y.~Sun, L.~Zheng, Y.~Yang, Q.~Tian, and S.~Wang, ``Beyond part models: Person
  retrieval with refined part pooling (and a strong convolutional baseline),''
  in \emph{Proceedings of the European Conference on Computer Vision (ECCV)},
  2018, pp. 501--518.

\bibitem{2020AsNet}
S.~Zhang, L.~Zhang, W.~L. Wang, and X.~Wu, ``Asnet: Asymmetrical network for
  learning rich features in person re-identification,'' \emph{IEEE Signal
  Processing Letters}, vol.~27, pp. 850--854, 2020.

\bibitem{chen2020pseudo}
S.~Chen, Z.~Fan, and J.~Yin, ``Pseudo label based on multiple clustering for
  unsupervised cross-domain person re-identification,'' \emph{IEEE Signal
  Processing Letters}, vol.~27, pp. 1460--1464, 2020.

\bibitem{2019Person}
Q.~Yang, A.~Wu, and W.~S. Zheng, ``Person re-identification by contour sketch
  under moderate clothing change,'' \emph{IEEE Transactions on Pattern Analysis
  and Machine Intelligence}, vol.~43, no.~6, pp. 2029--2046, 2019.

\bibitem{wan2020person}
F.~Wan, Y.~Wu, X.~Qian, Y.~Chen, and Y.~Fu, ``When person re-identification
  meets changing clothes,'' in \emph{Proceedings of the IEEE/CVF Conference on
  Computer Vision and Pattern Recognition Workshops (CVPRW)}, 2020, pp.
  3620--3628.

\bibitem{yu2020cocas}
S.~Yu, S.~Li, D.~Chen, R.~Zhao, J.~Yan, and Y.~Qiao, ``Cocas: A large-scale
  clothes changing person dataset for re-identification,'' in \emph{Proceedings
  of the IEEE/CVF Conference on Computer Vision and Pattern Recognition
  (CVPR)}, 2020, pp. 3400--3409.

\bibitem{fan2020learning}
L.~Fan, T.~Li, R.~Fang, R.~Hristov, Y.~Yuan, and D.~Katabi, ``Learning longterm
  representations for person re-identification using radio signals,'' in
  \emph{Proceedings of the IEEE/CVF Conference on Computer Vision and Pattern
  Recognition (CVPR)}, 2020, pp. 10\,699--10\,709.

\bibitem{2019Gait}
Z.~Ziyuan, T.~Luan, Y.~Xi, A.~Yousef, L.~Xiaoming, W.~Jian, and W.~Nanxin,
  ``Gait recognition via disentangled representation learning,'' in
  \emph{Proceedings of the IEEE/CVF Conference on Computer Vision and Pattern
  Recognition (CVPR)}, 2019, pp. 4710--4719.

\bibitem{2019EV_Gait}
W.~Yanxiang, D.~Bowen, S.~Yiran, W.~Kai, Z.~Guangrong, S.~Jianguo, and
  W.~Hongkai, ``Ev-gait: Event-based robust gait recognition using dynamic
  vision sensors,'' in \emph{Proceedings of the IEEE/CVF Conference on Computer
  Vision and Pattern Recognition (CVPR)}, 2020, pp. 6358--6367.

\bibitem{2020Gait_Chao}
F.~Chao, P.~Yunjie, C.~Chunshui, L.~Xu, H.~Saihui, C.~Jiannan, H.~Yongzhen,
  L.~Qing, and H.~Zhiqiang, ``Gaitpart: Temporal part-based model for gait
  recognition,'' in \emph{Proceedings of the IEEE/CVF Conference on Computer
  Vision and Pattern Recognition (CVPR)}, 2020, pp. 14\,213--14\,221.

\bibitem{2021Cloth_Changing}
J.~Xin, H.~Tianyu, Z.~Kecheng, Y.~Zhiheng, S.~Xu, H.~Zhen, F.~Ruoyu,
  H.~Jianqiang, H.~Xian-Sheng, and C.~Zhibo, ``Cloth-changing person
  re-identification from a single image with gait prediction and
  regularization,'' in \emph{arXiv preprint arXiv:2103.15537}, 2021.

\bibitem{zhang2019freeanchor}
X.~Zhang, F.~Wan, C.~Liu, R.~Ji, and Q.~Ye, ``Freeanchor: Learning to match
  anchors for visual object detection,'' in \emph{Advances in neural
  information processing systems (NeurIPS)}, 2019, pp. 147--155.

\bibitem{li2020self}
P.~Li, Y.~Xu, Y.~Wei, and Y.~Yang, ``Self-correction for human parsing,''
  \emph{IEEE Transactions on Pattern Analysis and Machine Intelligence}, 2020.

\bibitem{liang2015deep}
X.~Liang, S.~Liu, X.~Shen, J.~Yang, L.~Liu, J.~Dong, L.~Lin, and S.~Yan, ``Deep
  human parsing with active template regression,'' \emph{IEEE transactions on
  pattern analysis and machine intelligence}, vol.~37, no.~12, pp. 2402--2414,
  2015.

\bibitem{he2016deep}
K.~He, X.~Zhang, S.~Ren, and J.~Sun, ``Deep residual learning for image
  recognition,'' in \emph{Proceedings of the IEEE/CVF Conference on Computer
  Vision and Pattern Recognition (CVPR)}, 2016, pp. 770--778.

\bibitem{russakovsky2015imagenet}
O.~Russakovsky, J.~Deng, H.~Su, J.~Krause, S.~Satheesh, S.~Ma, Z.~Huang,
  A.~Karpathy, A.~Khosla, M.~S. Bernstein \emph{et~al.}, ``Imagenet large scale
  visual recognition challenge,'' \emph{International Journal of Computer
  Vision}, vol. 115, no.~3, pp. 211--252, 2015.

\bibitem{li2018harmonious}
W.~Li, X.~Zhu, and S.~Gong, ``Harmonious attention network for person
  re-identification,'' in \emph{Proceedings of the IEEE conference on computer
  vision and pattern recognition (CVPR)}, 2018, pp. 2285--2294.

\bibitem{Zhong2017Re}
Z.~Zhong, L.~Zheng, D.~Cao, and S.~Li, ``Re-ranking person re-identification
  with k-reciprocal encoding,'' in \emph{Proceedings of the IEEE/CVF Conference
  on Computer Vision and Pattern Recognition (CVPR)}, 2017, pp. 3652--3661.

\bibitem{wang2018learning}
G.~Wang, Y.~Yuan, X.~Chen, J.~Li, and X.~Zhou, ``Learning discriminative
  features with multiple granularities for person re-identification,'' in
  \emph{Proceedings of the 26th ACM international conference on Multimedia},
  2018, pp. 274--282.

\bibitem{fu2019horizontal}
Y.~Fu, Y.~Wei, Y.~Zhou, H.~Shi, G.~Huang, X.~Wang, Z.~Yao, and T.~S. Huang,
  ``Horizontal pyramid matching for person re-identification,'' in
  \emph{Proceedings of the AAAI Conference on Artificial Intelligence (AAAI)},
  vol.~33, no.~01, 2019, pp. 8295--8302.

\bibitem{2015Karen}
S.~Karen and Z.~Andrew, ``Very deep convolutional networks for large-scale
  image recognition,'' in \emph{International Conference on Learning
  Representations (ICLR)}, 2015.

\bibitem{wu2018exploit}
Y.~Wu, Y.~Lin, X.~Dong, Y.~Yan, W.~Ouyang, and Y.~Yang, ``Exploit the unknown
  gradually: One-shot video-based person re-identification by stepwise
  learning,'' in \emph{Proceedings of the IEEE Conference on Computer Vision
  and Pattern Recognition}, 2018, pp. 5177--5186.

\bibitem{luo2019bag}
H.~Luo, Y.~Gu, X.~Liao, S.~Lai, and W.~Jiang, ``Bag of tricks and a strong
  baseline for deep person re-identification,'' in \emph{CVPR Workshops}, 2019,
  pp. 1487--1495.

\bibitem{2016A}
W.~Yandong, Z.~Kaipeng, L.~Zhifeng, and Q.~Yu, ``A discriminative feature
  learning approach for deep face recognition,'' in \emph{Proceedings of the
  European Conference on Computer Vision (ECCV)}, 2016, pp. 499--515.

\bibitem{zhong2020random}
Z.~Zhong, L.~Zheng, G.~Kang, S.~Li, and Y.~Yang, ``Random erasing data
  augmentation.'' in \emph{Proceedings of the AAAI Conference on Artificial
  Intelligence (AAAI)}, 2020, pp. 13\,001--13\,008.

\bibitem{zhou2016learning}
B.~Zhou, A.~Khosla, A.~Lapedriza, A.~Oliva, and A.~Torralba, ``Learning deep
  features for discriminative localization,'' in \emph{Proceedings of the
  IEEE/CVF Conference on Computer Vision and Pattern Recognition (CVPR)}, 2016,
  pp. 2921--2929.

\end{thebibliography}

\end{document}